\documentclass[superscriptaddress,10pt,twocolumn,nofootinbib]{revtex4-2} 

\usepackage{mathtools}

\usepackage{amsmath,booktabs,longtable,array}

\usepackage[bottom]{footmisc}

\usepackage{hyperref}

\usepackage{enumitem}

\usepackage{booktabs,makecell,array}
\newcolumntype{P}[1]{>{\raggedright\arraybackslash}p{#1}}

\usepackage[most]{tcolorbox}

\usepackage{amsmath}  
\usepackage{amsfonts} 
\usepackage{graphicx} 
\usepackage{esvect}
\usepackage{mathtools}

\usepackage{xcolor}
\usepackage{appendix}
\usepackage{siunitx} 
\usepackage{mathrsfs}

\usepackage{placeins}

\usepackage[caption=false]{subfig} 

\usepackage{pbox}

\usepackage{mathtools}
\makeatletter
\newcommand{\vast}{\bBigg@{3}}
\newcommand{\Vast}{\bBigg@{3.5}}
\makeatother

\begin{document}


\title{Geometry-Agnostic Data-Driven Discovery of Extreme Scenarios}


\author{Angeline Hu}
\thanks{These authors contributed equally.}
\affiliation{Claremont McKenna College, Claremont, CA 91711, USA.}

\author{Ava K. Tse}
\thanks{These authors contributed equally.}
\affiliation{Scripps College, Claremont, CA 91711, USA.}

\author{Kaitlyn S. Yasumura}
\thanks{These authors contributed equally.}
\affiliation{Scripps College, Claremont, CA 91711, USA.}

\author{Luciana Sarabia}
\thanks{These authors contributed equally.}
\affiliation{Scripps College, Claremont, CA 91711, USA.}

\author{Malcolm Yeung}
\affiliation{Pitzer College, Claremont, CA 91711, USA.}

\author{Kavin K. Prasanna}
\affiliation{Claremont McKenna College, Claremont, CA 91711, USA.}

\author{Kai Wong}
\affiliation{Pitzer College, Claremont, CA 91711, USA.}

\author{William Hammond}
\affiliation{Pitzer College, Claremont, CA 91711, USA.}

\author{Ioannis G. Keverkidis}
\affiliation{Department of Chemical and Biomolecular Engineering, \\ John Hopkins University, Baltimore, MD 21218, USA}

\author{Shengkai Li}
\affiliation{Department of Physics, Princeton University, Princeton, NJ 08544, USA}

\author{Trung V. Phan}
\email{{tphan@natsci.claremont.edu}}
\affiliation{Department of Natural Sciences, Scripps and Pitzer Colleges, \\ Claremont Colleges Consortium, Claremont, CA 91711, USA}

\begin{abstract}
We introduce a two-stage \textit{data-driven} neural framework for discovering dynamically admissible extreme scenarios. Rather than assuming prior knowledge of the governing equations, the first stage infers the effective dynamical constraints directly from observational data. When the underlying physical system possesses suitable symmetries, the learned dynamics can be represented in a \textit{geometry-agnostic} form,
independent of any fixed grid, coordinate system, dimensionality, and intrinsic curvature. In the second stage, a \textit{physics-informed} neural-search uses this learned dynamics to uncover rare, high-impact coherent
structures that may not have appeared in the original observations. We demonstrate this framework through representative examples of increasing dimensionality and geometric complexity. Starting from a traditionally learned Duffing dynamics, we find a \textit{homoclinic pulse} as a plausible localized extreme scenario. Next, since the same equation also appears in the stationary nonlinear Schr\"odinger system, with the coordinate interpreted as space instead of time, this pulse becomes a \textit{standing soliton}. We then retrain the first stage using a geometry-agnostic neural representation, enabling the learned dynamics to be lifted to higher-dimensional space to obtain higher-dimensional solitary profile, known as Townes soliton. Finally, with a Liouville system, we demonstrate that dynamics learned from data on a $(1+1)$-dimensional Lorentzian quasi-manifold can be transferred to a $(2+0)$-dimensional Euclidean manifold to identify an \textit{instanton}. Our approach can be applied to other coherent spatiotemporal structures, including kinks, rogue waves, compactons, peakons, and potentially even singularities, with applications in risk analysis, nonlinear forecasting, and anomaly detection.
\end{abstract}

\date{\today}

\maketitle 

\section{Introduction}

Extreme events are rare, often localized, departures from typical behavior \cite{ghil2011extreme}; in nonlinear systems, they are not necessarily external anomalies, but can arise naturally from the governing dynamics themselves \cite{chowdhury2022extreme}. Examples include rogue waves in nonlinear wave systems \cite{onorato2013rogue,kharif2003physical}, abrupt transitions in fluids and climate \cite{dysthe2008oceanic,wang2020experimental}, focusing stress in engineered structures \cite{charalampidis2018phononic,muralidhar2025extreme}, epidemic spikes in complex interconnected networks \cite{barthelemy2005dynamical}, anomalous financial events \cite{yan2010financial,yan2011vector} (e.g. market crashes \cite{jenks2020hyperwave,el2026financial}), and coherent bursts in complex systems \cite{frolov2019statistical}. These events can dominate the risk profile \cite{sornette2009dragon,sornette2012dragon} and exert a disproportionate influence on observable expectation values \cite{coleman1979uses, schafer1998instantons} or long-time dynamical behavior \cite{weinan2004minimum,bovier2016metastability} of the system. Despite their importance, they are unlikely to appear in naturally sampled data, especially in high-dimensional systems governed by ordinary differential equations (ODEs) or infinite-dimensional systems governed by field-theoretic partial differential equations (PDEs). This creates a major challenge: the available observations may contain only typical behavior, while the most informative and consequential \textit{scenarios} remain unseen. Here, we use the term \textit{scenario} to emphasize that the target object has not been observed; rather, it is a dynamically plausible event or structure supported by the learned system. It is therefore valuable to develop methods that can infer plausible extreme scenarios directly from data.

\begin{figure*}[!htbp]
\includegraphics[width=\textwidth]{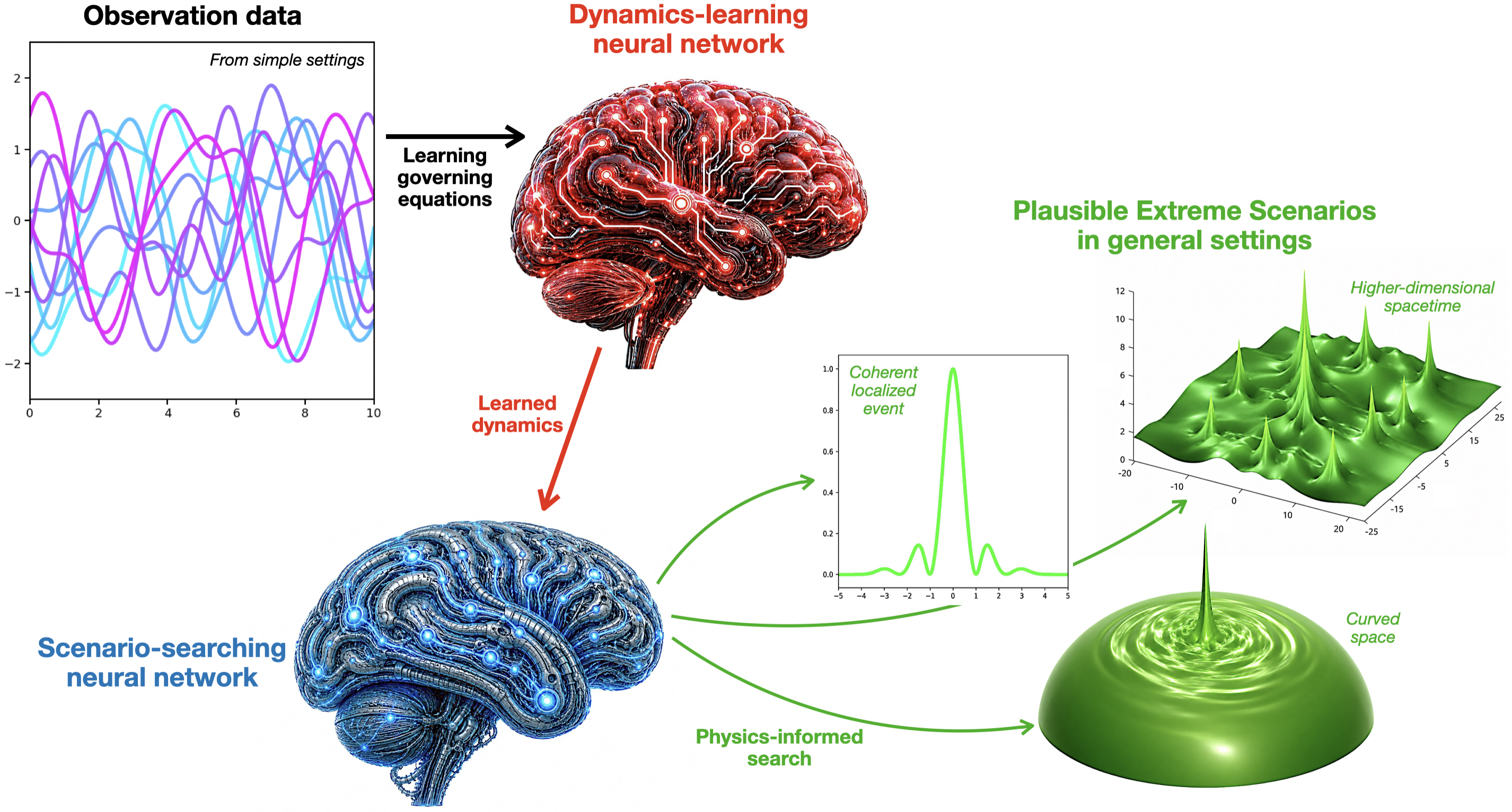}
\caption{\textbf{Overview of the data-driven two-stage neural framework for extremescenario discovery.} Observational data from simple or experimentally accessible settings are first used to train a dynamics-learning neural network that infers the underlying governing dynamics. The learned dynamics are then passed to a scenario-searching neural network, which performs a physics-informed search for dynamically admissible coherent structures. This second stage can identify plausible extreme scenarios that need not appear in the original observations, including localized events in higher-dimensional spacetime or in different geometric settings such as curved spaces.}
\label{fig01}
\end{figure*}

Recent advances in machine learning have created new tools for learning, solving, and forecasting nonlinear dynamical systems from data. Physics-informed neural network (PINN) uses automatic differentiation and differential-equation residuals to solve forward and inverse problems involving nonlinear PDEs \cite{raissi2019physics}. Related PINN-based methods have been used to reconstruct localized nonlinear-wave structures, including solitons, breathers, and rogue waves in integrable and non-integrable systems \cite{pu2021soliton,zhou2021deep}. More broadly, sparse-regression methods such as SINDy have shown that governing equations can sometimes be discovered directly from data when the dynamics are sparse in a suitable library of candidate terms \cite{brunton2016discovering}. Neural operators, including DeepONet and Fourier neural operators, have extended data-driven modeling from finite-dimensional input-output maps to operators between function spaces, enabling fast surrogate modeling for families of PDE solutions \cite{lu2021learning,li2020fourier}. In parallel, active learning strategies have been combined with neural operators to improve the discovery and forecasting of extreme events by selecting informative data in high-dimensional systems \cite{pickering2022discovering}. Structure-preserving neural networks further show that embedding conservation laws, Hamiltonian structure, geometric constraints, or thermodynamic consistency into learning architectures can improve physical plausibility and long-time stability \cite{greydanus2019hamiltonian,bertalan2019learning,hernandez2021structure,jin2022learning}. These developments form part of a substantial literature on the mechanisms, prediction, and statistical quantification of extreme events \cite{pickering2022discovering,farazmand2017variational}. Particularly relevant is the variational framework of Farazmand et al. \cite{farazmand2017variational}, which identifies extreme event precursors through constrained optimization with conditions ensuring their probabilistic feasibility. More recently, Thiel et al. \cite{thiel2026mechanism} used ensemble-derived information about local instability to improve rare-event statistics in neural emulators trained on limited data. Our work contributes to this broader effort by combining learned local dynamical constraints with a separate neural-search for prescribed classes of coherent structures, including transfer across geometric settings when the governing laws admit a suitable invariant representation. The present demonstrations concern the construction of candidate structures; their occurrence probabilities and statistical significance remain to be established. 

In this work, we introduce a two-stage formulation of extreme scenario discovery (see Fig. \ref{fig01}), starting from data. The first stage learns the underlying dynamics from observational data, while the second searches the learned dynamics for dynamically admissible coherent structures. For simplicity, we work with \textit{synthetic data}, focus on \textit{localized structures}, and implement the framework using simple multi-layer perceptron (MLP) neural networks with tanh-activation functions --- our aim is to demonstrate that the proposed approach can be implemented with minimal architectural complexity while still producing reasonable results. Perhaps most importantly, we demonstrate that our proposed formulation allows the first stage to be \textit{geometry-agnostic}, so that the learned dynamics are not tied to any particular grid, coordinate system, dimensionality, or geometric structure. Compared with the previous proposal on Riemannian manifolds \cite{phan2025towards}, we extend this idea to a broader class of Riemann-Cartan geometric-manifolds (including quasi-manifolds and torsionful geometries, in which the local structure depends not only on the metric but also on an independent affine connection). This flexibility enables data from different settings to be aggregated and then used to make predictions in completely new settings. This capability is especially important when the extreme scenario of interest lives in an experimentally inaccessible space. Such a situation arises naturally in many relativistic field-theoretic systems \cite{coleman1979uses}, where instanton solutions are formulated on an imaginary-time Euclidean manifold rather than on the real-time Lorentzian quasi-manifold representing physical spacetime.

\section{Computational Experiments}

We show proposed our data-driven neural framework through three examples that progress from a simple ODE setting to nonlinear PDEs with different dimensionalities and spacetime signatures\footnote{Throughout, we denote an $(p+q)$-dimensional relativistic system as one with $p$ spatial dimensions and $q$ time dimensions, and we use the metric signature convention in which spatial directions have positive sign and time directions have negative sign. This system possesses an O$(p,q)$ symmetry, corresponding to Lorentzian rotations and boosts, as well as time- and space-reversal symmetries.}:
\begin{itemize}
    \item \underline{Section \ref{sec:fullpipe}:} We consider a Duffing ODE \cite{kovacic2011duffing} to illustrate the complete pipeline: learning the dynamics from data and then searching the learned system for a \textit{homoclinic pulse} \cite{balmforth1995solitary}. We also examine a control case in which the learned ODE does not support such a solution (i.e. simple harmonic system), demonstrating how the search fails when the target structure is not dynamically 
    admissible. 
    \item \underline{Section \ref{sec:townes}:} We use the stationary nonlinear (SNL) Schrödinger equation \cite{sulem2007nonlinear} to show how learned dynamics can be lifted from a one-dimensional problem to a higher-dimensional spatial setting. The Duffing equation considered previously can be reinterpreted as a SNL Schr\"odinger equation in one-dimensional space, turning the localized pulse into a standing soliton profile. After retraining the first stage with a geometry-agnostic neural representation, the learned dynamics can be generalized to two spatial dimensions, or higher via the O$(p)$ symmetry\footnote{Here, $p$ denotes the number of spatial dimensions, and the O$(p)$ symmetry includes both rotations and spatial reversals.}, where the scenario-searching network identifies a self-focusing Townes profile \cite{chiao1964self}. This example demonstrates that data from a lower-dimensional setting can support the prediction of a coherent localized structure in a higher-dimensional space.
    \item \underline{Section \ref{sec:geoag}:} The third example explores how the two-stage formulation creates room for novelty at each stage of the framework. We study the Liouville \textit{relativistic} PDE \cite{nakayama2004liouville} to extend a previously developed geometry-agnostic learning framework on \textit{torsionless Riemannian manifolds} \cite{phan2025towards} to also include \textit{torsionful Lorentzian quasi-manifolds}, and then incorporate this extension into our extreme-scenario discovery pipeline. Dynamics learned from data on a $(1+1)$-dimensional spacetime are transferred to a $(2+0)$-dimensional Euclidean space (obtained via a temporal Wick rotation), where they serve as the residual constraint for the hunt of \textit{instanton} solution \cite{coleman1979uses}. This example shows that learned dynamics can be used beyond the geometric setting in which the data were collected, enabling coherent extreme scenarios to be discovered in spaces that may not be even experimentally observable.
\end{itemize}

The scenario-searching network represents a candidate solution: its inputs are temporal or spatial coordinates, and its output is the normalized field profile. Automatic differentiation provides the derivatives needed to evaluate the learned dynamical constraint at sampled collocation points. During this stage, the dynamics-learning network is held fixed, while the parameters of the scenario-searching network and the associated coordinate- and field-rescaling factors are optimized. The loss combines the learned dynamical residual with pinning and localization conditions that specify the class of structures sought. No labeled examples of the target extreme profile enter this loss. Thus, the search determines a profile within a prescribed class, together with its physical scales; it does not automatically select which class of extreme scenarios to investigate.

We describe the procedure used to generate the training data (for the dynamics-learning neural networks) and the training locations (for the scenario-searching neural networks) in Appendix \ref{app:data}. The MLP neural-network architectures used in the examples are summarized in Appendix \ref{app:NN}, and the training procedures for the individual networks are given in Appendix \ref{app:training}. In the following sections of the main manuscript, we focus on: 
\begin{enumerate}[label=(\roman*), leftmargin=*]
    \item the mathematical description of the dynamical systems of interest and the corresponding extreme scenarios,
    \item what are the inputs and outputs of the neural networks,
    \item the loss functions used for the neural networks and the associated trainable parameters, and
    \item the comparison between the neural-found and analytical solutions.
\end{enumerate}

\section{Full Pipeline: Searching for \\ a Duffing Homoclinic Pulse \label{sec:fullpipe}}

\begin{figure*}[!htpb]
\includegraphics[width=\textwidth]{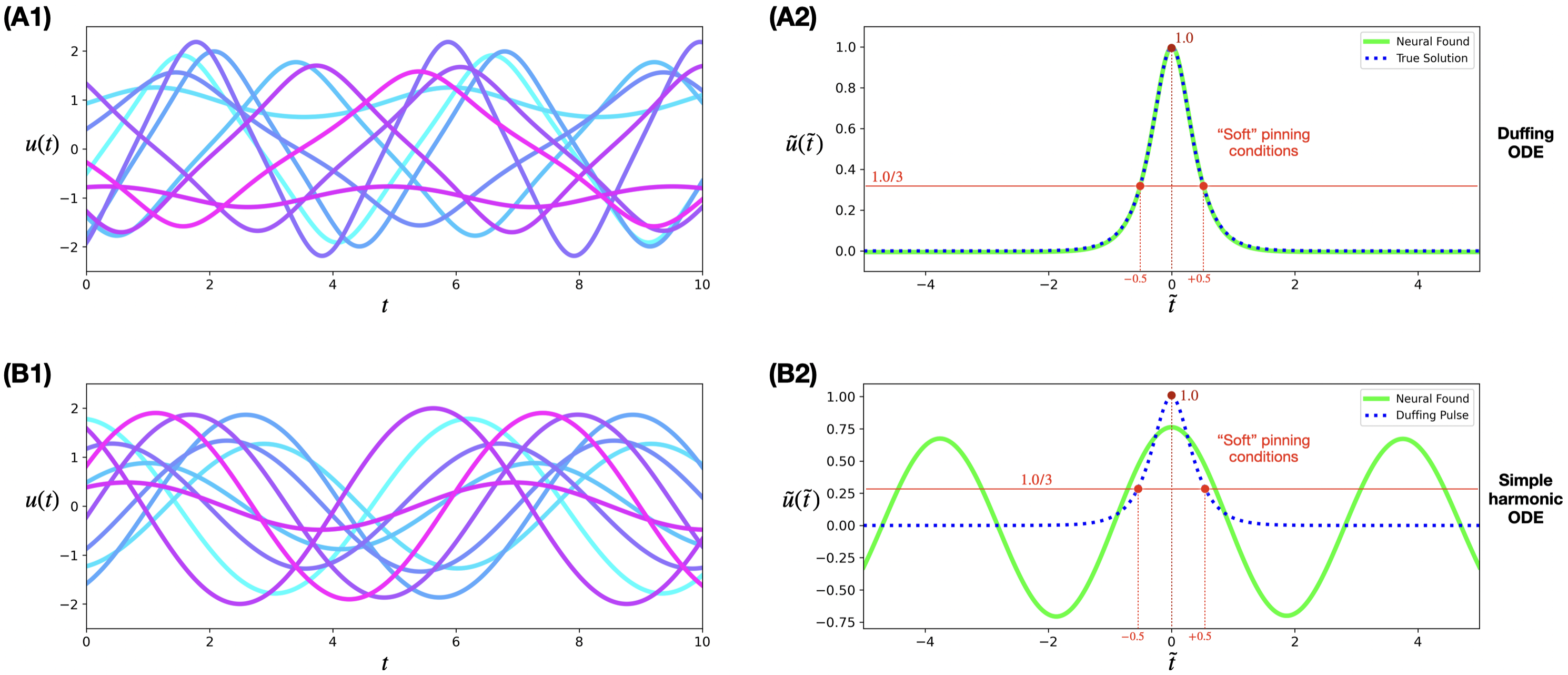}
\caption{\textbf{Full two-stage pipeline for learning dynamics and searching for a homoclinic pulse.} The comparison between the Duffing \textbf{(A1,A2)} and simple harmonic systems \textbf{(B1,B2)} provides a positive and negative test of the framework: the search should recover a pulse when the learned dynamics support one, and fail when they do not. \textbf{(A1,B1)} The representative trajectories of data used to train the dynamics-learning neural network in the first stage. \textbf{(A2,B2)} The result of the scenario-searching neural network in the second stage, with pinning conditions explicitly shown. For the Duffing system \textbf{(A2)}, the search recovers a localized homoclinic pulse in good agreement with the analytical solution. For the simple harmonic system \textbf{(B2)}, the same search constraints do not produce a dynamically admissible pulse.}
\label{fig02}
\end{figure*}

We use the first example to illustrate the full two-stage pipeline: the first neural network learns the underlying dynamics from synthetic trajectory data (see Fig. \ref{fig02}A1), and the second neural network searches the learned dynamics for a localized coherent structure. The trainings are done consecutively.

We consider a one-dimensional Duffing ODE:
\begin{equation}
    \ddot{u}(t) = u(t) - u^3(t) \ ,
    \label{eq:Duffing}
\end{equation}
where $t$ is the time-coordinate, $u(t)$ is the scalar state variable and the dot denotes differentiation with respect to time, i.e. $$\dot{u}(t) =  \frac{d}{dt} u(t) \ , \ \ddot{u}(t) = \frac{d^2}{dt^2} u(t)  \ .$$
A homoclinic pulse is a localized trajectory that leaves an equilibrium state, develops a large transient excursion, and then returns to the same equilibrium state. This Duffing equation admits homoclinic pulse solutions of the form:
\begin{equation}
    u(t) = \pm \sqrt{2}\,\mathrm{sech}(t-t_*) ,
    \label{eq:homo_pulse}
\end{equation}
in which $t_*$ denotes the center time of the pulse (can be any real-value), i.e. the time at which the solution reaches its maximum amplitude.
Thus, the scalar state is asymptotically $0$ in the far past $t \rightarrow -\infty$, rapidly forms a pronounced peak $\pm \sqrt{2}$ at time $t=t_*$, and then decays back to $0$ in the far future $t \rightarrow +\infty$. Physically, this type of solution is a clear representation of a drastic extreme event: the system appears nearly silent before the event, suddenly produces a large-amplitude localized pulse, and then becomes silent again.

\underline{For the first stage,} following Appendix \ref{app:data_Duffing}, we generate a synthetic training dataset
$\{u_j,\dot{u}_j,\ddot{u}_j,...\}_{j=1}^{M_1^{(D)}}$ from randomly-selected trajectories of
Eq. \eqref{eq:Duffing}. Since the underlying dynamics are not known \textit{a priori}, we assume a broad class of local dynamical laws by retaining all candidate terms up to a prescribed derivative order $\delta$, which we choose to be $\delta=2$. We then train the neural parameters of the
dynamics-learning neural network, denoted by $\mathbb{N}_1^{(D)}$, which
takes two-inputs $(u_j,\dot{u}_j)$ and returns one-output
$\mathbb{N}_1^{(D)}(u_j,\dot{u}_j)$. A visualization of this neural network can be found in Fig. \ref{figS01}A1. This training is done by minimizing
the squared-deviation loss function:
$$\mathcal{L}_{\mathbb{N}_1^{(D)}} = \bar{E}\left\{ \left[ \mathbb{N}_1^{(D)}(u_j,\dot{u}_j) - \ddot{u}_j \right]^2 \right\}_{\forall j} \ , $$
where $\bar{E}\{ \hat{X} \}_{\mathcal{C}}$ denotes the averaging operation applied to the expression $\hat{X}$ over all data points satisfying the clause $\mathcal{C}$. After training, $\mathbb{N}_1^{(D)}$ encodes the data-driven neural-learned dynamics of the Duffing system, i.e.
\begin{equation}
    \mathbb{N}_1^{(D)}(u,\dot{u}) \approx \ddot{u} \ .
    \label{eq:Duffing_learned}
\end{equation}

\underline{For the second stage,} our main goal is to train a scenario-searching neural network, denoted as $\mathbb{N}_2^{(D)}$, so that with it we can identify a homoclinical pulse solution of Eq. \eqref{eq:Duffing}. Given that the pulse can occur at any time; thus, without loss of generality, we can gauge away this freedom by requiring that we only search for a solution centered exactly at $t=0$. Because the maximum amplitude and characteristic width of the pulse are not known
\textit{a priori}, we do not search for the solution $u(t)$
directly. Instead, we use the neural model $\mathbb{N}_2^{(D)}$ to approximate a \textit{normalized} pulse
profile $\tilde{u}(\tilde{t})$ in a \textit{normalized} time coordinate $\tilde{t}$, i.e. 
\begin{equation}
    \mathbb{N}_2^{(D)}(\tilde{t}) \approx \tilde{u}(\tilde{t}) \ .
    \label{eq:norm_homo_learned}
\end{equation}
A visualization of this neural network can be found in Fig. \ref{figS01}B1.

In this \textit{normalized} representation, the pulse center is fixed at $\tilde{t}=0$ and the peak amplitude is prescribed by imposing
\begin{equation}
    \tilde{u}(0)=1.0 \ .
    \label{eq:Duffing_peak_pinning}
\end{equation}
Additionally, we impose
shoulder conditions at $\tilde{t}=\pm 0.5$, such that it is lower than the peak by a chosen factor $\eta=3$:
\begin{equation}
    \tilde{u}(\pm 0.5)=\tilde{u}(0)/\eta = 1.0/3\ ,
    \label{eq:Duffing_shoulder}
\end{equation}
and the temporal localization requirement that such that far away amplitude stays small:
\begin{equation}
    \left| \tilde{u}(\tilde{t})\Big|_{|\tilde{t}|<0.5} \right| <\tilde{u}(0)/\eta = 1.0/3 \ ,
    \label{eq:Duffing_localization}
\end{equation}
The homoclinical pulse solution is then recovered through the rescaling:
\begin{equation}
    t = T^{(D)}\tilde{t},
    \qquad
    u(t) = U^{(D)}\tilde{u}(\tilde{t}),
    \label{eq:TU_Duffing}
\end{equation}
Thus, we need to introduce two more parameters, $\mathbb{T}^{(D)}$ and $\mathbb{U}^{(D)}$, to approximate $T^{(D)}$ and $U^{(D)}$, i.e.
\begin{equation}
    \mathbb{T}^{(D)} \approx T^{(D)} \ , \ \mathbb{U}^{(D)} \approx U^{(D)} \ .
    \label{eq:Duffing_TU_learned}
\end{equation}
These parameters should be trained simultaneously with the neural parameters of $\mathbb{N}_2^{(D)}$, so that while $\mathbb{N}_2^{(D)}$ learns the normalized pulse shape, $\mathbb{T}^{(D)}$ and $\mathbb{U}^{(D)}$ learn how this normalized profile should be embedded into the neural-learned Duffing dynamics $\mathbb{N}_1^{(D)}$, encoded by the constraint:
\begin{equation}
\begin{split}
    \mathbb{N}_1^{(D)}\left(U^{(D)} \mathbb{N}_2^{(D)}(\tilde{t}) ,\frac{U^{(D)}}{T^{(D)}} \frac{d}{d\tilde{t}}\mathbb{N}_2^{(D)}(\tilde{t}) \right) & 
    \\
    - \frac{U^{(D)}}{{T^{(D)}}^2} \frac{d^2}{d\tilde{t}^2} \mathbb{N}_2^{(D)}(\tilde{t}) = 0 & \ .
    \label{eq:Duffing_constraint}
\end{split}
\end{equation}
The required derivatives of the neural network $\mathbb{N}_2^{(D)}$
with respect to $\tilde{t}$ can be obtained by auto-differentiation. As a sanity check, Eq. \eqref{eq:Duffing_constraint} follows exactly from Eq. \eqref{eq:TU_Duffing} when the approximations in Eq. \eqref{eq:Duffing_learned}, Eq.\eqref{eq:norm_homo_learned}, Eq. \eqref{eq:Duffing_TU_learned} are all exact.

To train $\mathbb{N}_2^{(D)}$, $\mathbb{T}^{(D)}$, and $\mathbb{U}^{(D)}$, we create a set of training locations $\{t_j\}^{M^{(D)}_2}_{j=1}$, as follows from Appendix \ref{app:data_Duffing}. We then minimize a total loss
function: $$\mathcal{L}_{\mathbb{N}_2^{(D)},\mathbb{T}^{(D)},\mathbb{U}^{(D)}} = \sum^5_{j = 1} c^{(D)}_j \mathcal{L}^{(D)}_j \ , $$ 
where $c_{\alpha}^{(D)}>0$ are weighting coefficients. The five loss functions $\{ \mathcal{L}^{(D)}_j \}^5_{j=1}$
consist of:
\begin{itemize}
    \item The ``soft'' pinning condition Eq.  for the peak amplitude, i.e. \eqref{eq:Duffing_peak_pinning}:
    $$ \mathcal{L}^{(D)}_1 = \left[ \mathbb{N}_2^{(D)}(0)- 1.0 \right]^2 \ .$$ 
    \item The ``soft'' pinning conditions for the shoulders' values, i.e. Eq. \eqref{eq:Duffing_shoulder}:
    \begin{equation*}
    \begin{split}
        \mathcal{L}^{(D)}_2 &= \left[ \mathbb{N}_2^{(D)}(+0.5)- 1.0/3 \right]^2 \ , 
    \\
    \mathcal{L}^{(D)}_3 &= \left[ \mathbb{N}_2^{(D)}(-0.5)- 1.0/3 \right]^2 \ .
    \end{split}
    \end{equation*}
    \item The localization requirement, i.e. Eq. \eqref{eq:Duffing_localization}:
    $$\mathcal{L}^{(D)}_4 = \bar{E}\left\{ \left[ \text{ReLU}\left(|\mathbb{N}_2^{(D)}(\tilde{t}_j)|-1.0/3\right) \right]^2 \right\}_{|\tilde{t}_j| > 0.5} \ ,$$
    in which $\text{ReLU}(\zeta)$ is the rectified linear unit function, i.e. $\text{ReLU}(\zeta) = \max(0,\zeta)$.
    \item The neural-learned Duffing dynamics constraint, i.e. Eq. \eqref{eq:Duffing_constraint}:
    \begin{equation}
        \mathcal{L}^{(D)}_5 = \bar{E}\left\{ \left[ \hat{X}^{(D)}_1(\tilde{t}_j)-\hat{X}^{(D)}_2(\tilde{t}_j) \right]^2 \right\}_{\forall j} \ ,
    \label{eq:Duffing_dyn_constraint}
    \end{equation}
    in which the expressions are:
    \begin{equation*}
        \begin{split}
            \hat{X}^{(D)}_1(\tilde{t}_j) = \mathbb{N}_1^{(D)}\left(\mathbb{U}^{(D)} \mathbb{N}_2^{(D)}(\tilde{t}_j) ,\frac{\mathbb{U}^{(D)}}{\mathbb{T}^{(D)}} \frac{d}{d\tilde{t}}\mathbb{N}_2^{(D)}(\tilde{t}_j) \right) & \ ,
            \\
            \hat{X}^{(D)}_2(\tilde{t}_j) = \frac{\mathbb{U}^{(D)}}{{\mathbb{T}^{(D)}}^2} \frac{d^2}{d\tilde{t}^2} \mathbb{N}_2^{(D)}(\tilde{t}_j) & \ .
        \end{split}
    \end{equation*}
\end{itemize} 
Note that throughout the entire second stage, the neural model $\mathbb{N}_1^{(D)}$ is kept fixed. The weights are chosen to be $(c_1^{(D)},c_2^{(D)},c_3^{(D)},c_4^{(D)},c_5^{(D)})=(1,1,1,1,25)$.

After training, we compare the neural-found normalized pulse
$\mathbb{N}_2^{(D)}(\tilde{t})$ with the exact normalized homoclinic pulse
solution $\tilde{u}(\tilde{t})$, finding excellent agreement --- see
Fig. \ref{fig02}A2. Inside the training domain $\tilde{t} \in [-\tau,+\tau]$ where $\tau = 5.0$, we have found the relative root-mean-square ($L^2$-norm) error to be very low:
\begin{equation*}
    \frac{\bar{E}\left\{\left[ \mathbb{N}_2^{(D)}(\tilde{t}_j)     -  \tilde{u}(\tilde{t}_j) \right]^2\right\}_{\forall j}}{\bar{E} \left\{ \left[ \tilde{u}(\tilde{t}_j)\right]^2 \right\}_{\forall j}} \approx 0.05\% \ .
\end{equation*}
The neural-found rescaling factors are $\mathbb{T}^{(D)} \approx 3.42$ and $\mathbb{U}^{(D)} \approx 1.42$, respectively, which are also in close agreement with the corresponding exact values $T^{(D)} \approx 3.52$ and $U^{(D)} \approx 1.41$. Overall, this example provides a minimal proof of concept; agreement in the normalized profile and rescaling factors shows that the method captures both the event shape and its physical amplitude and time scale with high precision.

A similar neural-search (every step is the same as above) can also be applied to a dynamical system that intrinsically lacks a homoclinic pulse, such as the simple harmonic oscillator: 
\begin{equation*}
    \ddot{u}(t) =  -u(t) \ .
\end{equation*}
In this case, the dynamics-learning network is trained on randomly generated trajectories (see Fig. \ref{fig02}B1), and the scenario-searching network is then applied in the same way. The search fails to find a valid pulse-like solution, instead producing deviations from the learned dynamics (see Fig. \ref{fig02}B2). It is important to note that, from the trajectories alone (see Fig. \ref{fig02}A1 and B1), it is not obvious what kinds of behavior can emerge, let alone whether either system can support a homoclinic pulse.

\section{Searching for Extreme Scenarios \\ beyond Training Geometries}

In physical systems, the governing dynamics often respect underlying symmetries, such as rotational symmetry in classical space or Lorentz rotations and boosts in relativistic spacetime \cite{landau1975classical}. Similar symmetry structures can also emerge at low-energy limit of gapless condensed matter systems \cite{borgnia2015quasi,phan2013ballistic}. These symmetries strongly constrain the possible features that can appear in the dynamical laws. In many cases, this makes it possible to learn the dynamics in a coordinate- and dimension-agnostic form, independent of the specific intrinsic curvature of the underlying space. Experimentally, this is extremely useful because one can train or infer the dynamics in a simple setting, while still using the learned model to predict behavior in other geometries, dimensions, or coordinate systems. A central advantage of our data-driven search framework is therefore that, when the dynamics-learning procedure is formulated geometry-agnostically, the governing constraints learned in one setting can be applied across many geometric settings, allowing the framework to discover extreme scenarios even beyond where the observational data were collected.

We explain geometry-agnostic learning in Section \ref{sec:geoag_exp} in a more general form than previously proposed \cite{phan2025towards}. We then show how this framework can predict extreme scenarios in settings beyond the training data: first for finding soliton in higher dimension in Section \ref{sec:townes}, and then for instanton in different geometric structure (e.g. from a quasi-manifold to a manifold) in Section \ref{sec:geoag}.

\subsection{Geometry-Agnostic Learning \label{sec:geoag_exp}}

Geometry-agnostic learning means learning a physical rule in a form that is not tied to the particular geometry in which the data were measured. Rather than learning derivatives that only make sense in a specific coordinate system, the model learns from invariant geometric quantities. For scalar-field dynamics, these may include the field value, the Laplacian, and inner products of gradients. Because these quantities are defined intrinsically, the same learned law can be applied on a line, a plane, a curved surface, or a higher-dimensional space. In this sense, for systems with suitable symmetries, such representations are not only coordinate-independent but also trans-dimensional: the same feature set is well defined and can be evaluated consistently across spaces of different dimensions. 

In \cite{phan2025towards}, this idea was developed for scalar fields on a torsionless Riemannian manifold $(\mathcal{M},g)$, where $\mathcal{M}$ denotes the underlying space equipped with a metric structure $g$, under the assumption that the system possesses a local O$(p)$ symmetry. However, the same perspective can be extended to torsionful Riemann-Cartan manifolds $(\mathcal{M},g,\nabla_\Gamma )$, where the covariant derivative $\nabla_\Gamma$ is equipped with a affine connection $\Gamma$ that is not necessarily determined solely by the metric, as well as to quasi-manifolds.

For concreteness, consider a collection of $M$ scalar fields, $\{u_a\}_{a=1}^{M}$. The dynamical features used by the learning model can then be organized according to their jet-space order, or equivalently their derivative order. This derivative order also provides a natural hierarchy of importance, analogous to power counting in effective field theory, where low-resolution or long-wavelength dynamics are typically dominated by lower-derivative terms. For example, one may include all scalar invariant features up to second order (counting derivatives):
\begin{itemize}
    \item The field values $\{u_a\}_{a=1}^{M}$. 
    
    ($0$th-order features)
    \item $\{\langle \nabla_\Gamma u_a , \nabla_\Gamma u_b \rangle_g \}_{a,b=1}^{M}$ 
    and $\{\nabla_\Gamma^2 u_a\}_{a=1}^{M}$.
    
    ($2$nd-order features)
\end{itemize}
Note that there is no $1$st-order feature. The second-order features are the gradient inner products and the Laplacians, which can be written explicitly in tensor form as:
\begin{enumerate}
    \item The gradient inner products:
    \begin{equation}
        \langle \nabla_\Gamma u_a , \nabla_\Gamma u_b \rangle_g = g^{\mu\nu} \partial_\mu u_a \partial_\nu u_b \ ,
    \end{equation}
    where $g^{\mu\nu}$ is the inverse of the symmetric metric tensor $g_{\mu\nu}$ (i.e. $g^{\mu\nu} g_{\nu\rho}=1$ only when $\mu=\rho$, and $g_{\mu\nu}=g_{\nu\mu}$). Here, we have used the Einstein's repeated-index summation convention.
    \item The field's Bochner Laplacians:
    \begin{equation}
        \nabla_\Gamma^2 u_a  = |g|^{-1/2} \partial_\mu \left( |g|^{1/2} g^{\mu\nu} \partial_\nu u_a \right) - g^{\mu\nu} C^{\lambda}_{\mu\nu} \partial_\lambda u_a \ ,
    \end{equation}
    where $|g|$ is the determinant of the metric tensor $g_{\mu\nu}$, and $C^\lambda_{\mu\nu}$ is the distorsion tensor (which is, in general, does not have any symmetry) \cite{nakahara2018geometry}. On any torsion-free manifold or quasi-manifold, this tensor vanishes, and Bochner Laplacian reduces to the Laplace–Beltrami operator when acting on scalar fields. In Euclidean flat space it further reduces to the Laplace operator. 
\end{enumerate}
We can then formulated the dynamical laws either implicitly or explicitly, i.e. via $M$ constraints:
\begin{equation}
    \mathcal{F}_c  \left( \{u_a\}_{a=1}^{M}, \{\langle \nabla_\Gamma u_a , \nabla_\Gamma u_b \rangle \}_{a,b=1}^{M} \right) = \nabla^2_\Gamma u_c  \ ,
    \label{eq:free}
\end{equation}
which is exactly identical to the form of the neural network model used in Eq. \eqref{eq:Duffing_learned} for learning the dynamics, but also geometric-agnostic. We utilize this in Sections \ref{sec:townes} and \ref{sec:geoag}.

There is, however, an important caveat. On a manifold, the gradient inner products are \textit{sign-definite}: positive-definite for a positive metric signature, or negative-definite for the opposite signature. On a quasi-manifold with mixed signature, by contrast, these inner products can take either sign. Therefore, a model trained only on data that lives on a manifold may not, in general, extrapolate reliably to a quasi-manifold. The reverse direction, however, is always possible: training on a quasi-manifold can expose the model to both sign sectors, allowing the learned dynamics to be evaluated on definite-signature manifolds as special cases. Indeed, this is the situation we encounter in Section \ref{sec:geoag}, after performing a temporal Wick rotation \cite{wick1954properties}.

\subsection{Searching for a Townes Soliton \label{sec:townes}}

\begin{figure}[!htbp]
\includegraphics[width=0.45\textwidth]{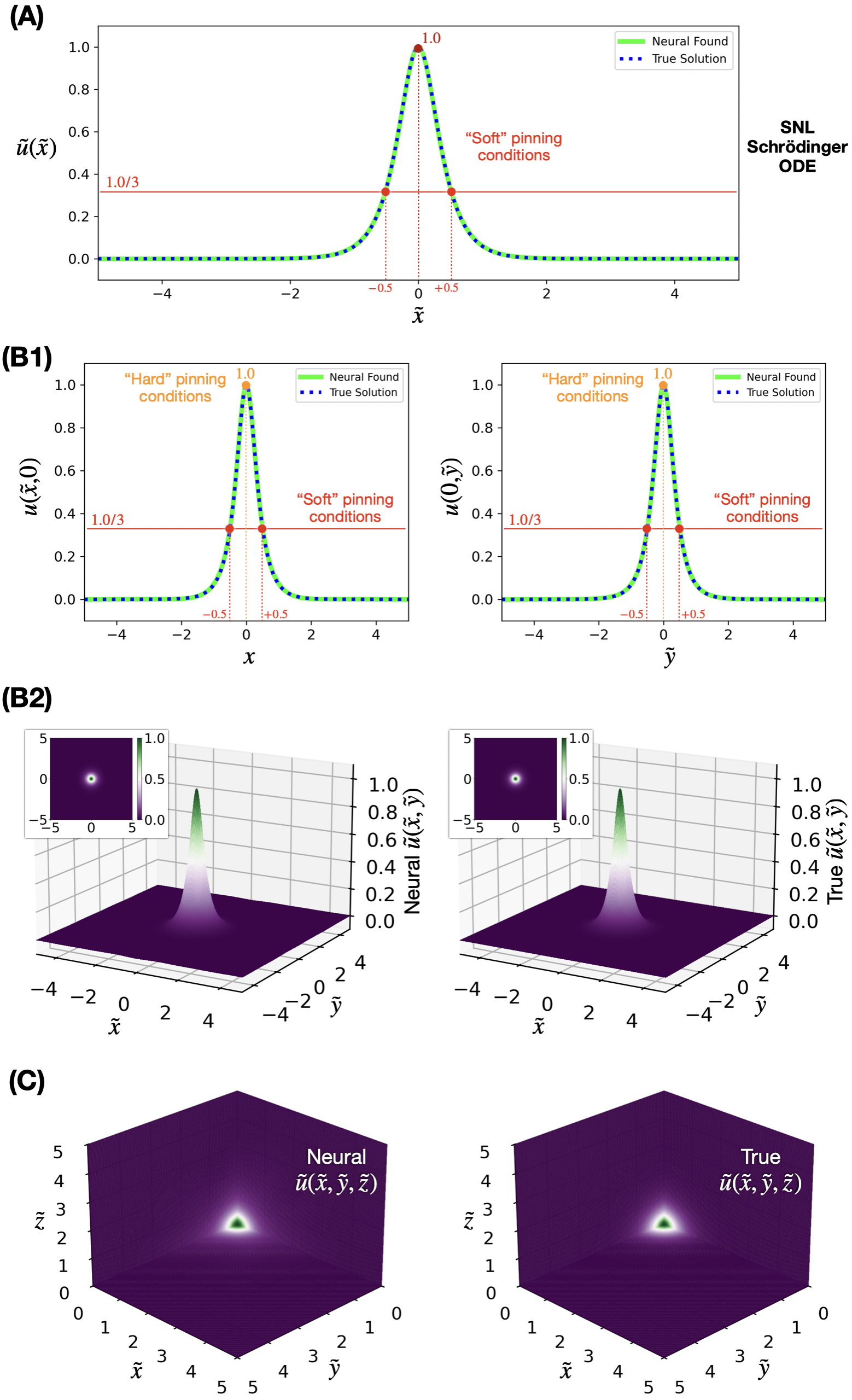}
\caption{\textbf{Geometry-agnostic neural-search result for stationary soliton profiles in the nonlinear Schr\"odinger system.} Those solutions are known as the Townes solitons. \textbf{(A)} The scenario-searching neural network recovers the one-dimensional stationary soliton profile, equivalent to the Duffing homoclinic pulse with the coordinate interpreted as space.
\textbf{(B1)} Cross sections of the two-dimensional neural-found profile along the $\tilde{x}$ and $\tilde{y}$.
\textbf{(B2)} Surface plots of the neural-found normalized Townes soliton and the high-precision numerical reference. \textbf{(C)} Three orthogonal cross sections of the three-dimensional soliton profile in the positive octant.}
\label{fig03}
\end{figure}

The Duffing system considered in Section \ref{sec:fullpipe} arises naturally as a stationary reduction of the nonlinear Schr\"odinger equation. To demonstrate this connection, we begin with the nonlinear Schr\"odinger PDE:
\begin{equation}
    \partial_t \Psi(\vec{r},t) + \vec{\nabla}^2 \Psi(\vec{r},t) + |\Psi(\vec{r},t)|^2 \Psi(\vec{r},t) = 0 \ , 
\label{eq:nonlin_Sch_full}
\end{equation}
where $\Psi(\vec{r},t)$ denotes a complex-valued wave field, $\vec{r}$ is the spatial coordinate, $t$ is time, and $\vec{\nabla}^2$ is the Laplace operator. 

We then consider stationary solutions of the form:
\begin{equation*}
    \Psi(\vec{r},t) = \exp(-i\omega t) u(\vec{r},t) \ ,
\end{equation*}
in which $u(\vec{r})$ is a time-independent real-valued field and $\omega$ is a real frequency parameter. Substituting this ansatz into Eq. \eqref{eq:nonlin_Sch_full} and factoring out the common oscillatory phase $\exp(-i\omega t)$ yields:
\begin{equation*}
    \vec{\nabla}^2 u(\vec{r}) = \omega u(\vec{r}) - u^3(\vec{r}) \ .
\end{equation*}
Up to the rescales $u \rightarrow \omega^{1/2} u$ and $x \rightarrow \omega^{-1/2} x$, we obtain the stationary nonlinear (SNL) Schr\"odinger equation:
\begin{equation}
    \vec{\nabla}^2 u(\vec{r}) = u(\vec{r}) - u^3(\vec{r}) \ .
\label{eq:SNL_Sch_full}
\end{equation}
In one-dimensional space $\vec{r}\equiv x$, this becomes the Duffing ODE, i.e. Eq. \eqref{eq:Duffing}, but with space-coordinate $x$ instead of time $t$:
$$u''(x) = u(x)-u^3(x) \ ; $$
here, the prime denotes differentiation with respect
to space, i.e.
$$u'(x) =  \frac{d}{dx} u(x) \ , \ u''(x) = \frac{d^2}{dx^2} u(x)  \ .$$
This equation governs the stationary configurations of the nonlinear Schr\"odinger system in one-dimensional space.

Given many of such configurations, similar to the first stage -- the dynamics-learning stage -- in Section \ref{sec:fullpipe}, we can train a neural network $\mathbb{N}^{(S_1)}_1$ to learn the dynamical constraint of the one-dimensional SNL Schr\"odinger. Note that, the superscript $^{(S_1)}$ here is to indicate that we are dealing with the one-dimensional SNL Schr\"odinger system. To make the learned dynamics geometry-agnostic, however, we should not use the coordinate-dependent form Eq. \eqref{eq:Duffing_learned} but the form proposed in Eq. \eqref{eq:free}, i.e.
\begin{equation}
    \mathbb{N}^{(S_1)}_1 ( u,{u'}^2 ) \approx u'' \ ,
    \label{eq:S1_train}
\end{equation}
where we have calculated in a flat Euclidean one-dimensional space that
$$\langle \nabla_\Gamma u,\nabla_\Gamma u \rangle = {u'}^2 \ , \ \nabla_\Gamma^2 u = u'' \ . $$
A visualization of this neural network can be found in Fig. \ref{figS01}A2. After training the network to a reasonably low loss, we freeze $\mathbb{N}^{(S_1)}_1$ and use it as the learned geometry-agnostic dynamical law in the second-stage search. 

\subsubsection{Sanity Check in 1D}

Before moving on, let us perform a sanity check. We can also find a pulse-like solution for $u(x)$, which should match the Duffing homoclinic pulse, as expressed by Eq. \eqref{eq:homo_pulse}, after replacing $t$ by $x$. In this context, this solution is the one-dimensional stationary soliton profile of the nonlinear Schr\"odinger equation. The neural-search procedure is identical to the second stage in Section \ref{sec:fullpipe}, except that the coordinate and its corresponding trainable rescaling parameter now acts on space rather than time. We denote the coordinate by $x$ this parameter by $R$, simply to emphasize its spatial interpretation; mathematically, it plays the same role as the temporal coordinate $t$ and the rescaling parameter $T$ used in the Duffing example. The loss contribution from the dynamical constraint, i.e. Eq. \eqref{eq:Duffing_dyn_constraint}, is then modified accordingly:
    \begin{equation}
        \mathcal{L}^{(S_1)}_5 = \bar{E}\left\{ \left[ \hat{X}^{(S_1)}_1(\tilde{x}_j)-\hat{X}^{(S_1)}_2(\tilde{x}_j) \right]^2 \right\}_{\forall j} \ ,
    \label{eq:SNL_Schr_1D_dyn_constraint}
    \end{equation}
    in which the expressions are:
    \begin{equation*}
        \begin{split}
            \hat{X}^{(S_1)}_1(\tilde{x}_j) = \mathbb{N}_1^{(S_1)}\left(\mathbb{U}^{(S_1)} \mathbb{N}_2^{(D)}(\tilde{x}_j) ,\left[ \frac{\mathbb{U}^{(S_1)}}{\mathbb{R}^{(S_1)}} \frac{d}{d\tilde{x}}\mathbb{N}_2^{(D)}(\tilde{x}_j) \right]^2 \right) & \ ,
            \\
            \hat{X}^{(D)}_2(\tilde{x}_j) = \frac{\mathbb{U}^{(D)}}{{\mathbb{R}^{(D)}}^2} \frac{d^2}{d\tilde{x}^2} \mathbb{N}_2^{(D)}(\tilde{x}_j) & \ .
        \end{split}
    \end{equation*}

We find that neural-search result for the normalized pulse
$\mathbb{N}_2^{(S_1)}(\tilde{x})$\footnote{A visualization of this neural network can be found in Fig. \ref{figS01}B1.} agrees with the exact normalized homoclinic pulse
solution $\tilde{u}(\tilde{x})$ --- see
Fig. \ref{fig03}A. Inside the training domain $\tilde{x} \in [-\lambda,+\lambda]$ where $\lambda = 5.0$, we have found the relative root-mean-square ($L^2$-norm) error to be extremely low:
\begin{equation*}
    \frac{\bar{E}\left\{\left[ \mathbb{N}_2^{(S_1)}(\tilde{x}_j)     -  \tilde{u}(\tilde{x}_j) \right]^2\right\}_{\forall j}}{\bar{E} \left\{ \left[ \tilde{u}(\tilde{x}_j)\right]^2 \right\}_{\forall j}} \approx 0.003\% \ ,
\end{equation*}
and rescaling factors are $\mathbb{R}^{(S_1)} \approx 3.62$ and $\mathbb{U}^{(S_2)} \approx 1.41$, respectively, which are close to their corresponding exact values $R^{(S_1)} \approx 3.52$ and $U^{(S_1)} \approx 1.41$. 

\subsubsection{neural-search in 2D \label{sec:search_SNL_Schr}}

We have trained $\mathbb{N}^{(S)}_1$ using only observational data taken from one-dimensional space. We now apply the same learned dynamics in two spatial dimensions $\vec{r}=(x,y)$, where we search for the corresponding Townes soliton profile in the SNL Schr\"odinger dynamics governed by Eq. \eqref{eq:SNL_Sch_full}. To do so, we must calculate the expression for the geometry-agnostic features in the new flat Euclidean two-dimensional space:
\begin{equation}
\begin{split}
    \langle \nabla_\Gamma u,\nabla_\Gamma u \rangle &= (\partial_x u)^2 + (\partial_y u)^2 \ ,
    \\
    \nabla_\Gamma^2 u &= \partial_x^2 u + \partial_y^2 u \ .
\end{split}
\label{eq:2D_feature}
\end{equation}
This step (and the calculated expressions) appears simple here only because we have chosen a minimal example to illustrate the idea. More generally, the target space may be curved, torsionful, or arbitrarily high-dimensional. The framework is unchanged: one simply recomputes the geometry-agnostic features using the appropriate geometric operators of the new space.

With the features in Eq. \eqref{eq:2D_feature}, we can now do the neural-search with a set of training locations $\{(\tilde{x}_j,\tilde{y}_j)\}^{M_2^{(S_2)}}_{j=1}$, which can be selected as explained in Appendix \ref{app:data_Schr}. Note that, the superscript $^{(S_2)}$ here is to indicate that we are dealing with the two-dimensional SNL Schr\"odinger system. Beside the neural parameters of $\mathbb{N}_2^{(S_2)}(\tilde{x},\tilde{y})$\footnote{A visualization of this neural network can be found in Fig. \ref{figS01}B2.}, which after a \textit{successful} search should approximate the \textit{normalized} Townes soliton $\tilde{u}(\tilde{x},\tilde{y})$, we need two more trainable parameters, $\mathbb{R}^{(S_2)}$ and $\mathbb{U}^{(S_2)}$, to approximate the spatial- and field-value-rescaling factors $R^{(S_2)}$ and $U^{(S_2)}$. It is important to note that no closed-form formula is available for this profile; however, given the known governing PDE, a symmetric solution of $\tilde{u}(\tilde{x},\tilde{y})$ can still be computed numerically to high precision (see Appendix \ref{app:2D_soliton_num}).

For this neural task, which requires searching over a two-dimensional profile, it is useful to impose as many ``hard'' constraints as possible. For example, if we seek a symmetric soliton centered at the origin, i.e.
$$\mathbb{N}^{(S_2)}_2(\tilde{x},\tilde{y}) = \mathbb{N}^{(S_2)}_2(\pm \tilde{x},\mp \tilde{y}) \ ,  $$
and want to pin its value to be exactly one there, i.e.
$$\mathbb{N}^{(S_2)}_2(0,0) = 1 \ ,  $$
then we can always construct it by symmetrizing and normalizing the output of an unconstrained neural network $\breve{\mathbb{N}}^{(S_2)}_2(\tilde{x},\tilde{y})$ as follows:
\begin{equation*}
\begin{split}
    \mathbb{N}^{(S_2)}_2(\tilde{x},\tilde{y}) = &\frac14 \Big[ \breve{\mathbb{N}}^{(S_2)}_2(\tilde{x},\tilde{y}) + \breve{\mathbb{N}}^{(S_2)}_2(-\tilde{x},-\tilde{y})
    \\
    & \ \ \ \ + \breve{\mathbb{N}}^{(S_2)}_2(-\tilde{x},\tilde{y}) + \breve{\mathbb{N}}^{(S_2)}_2(\tilde{x},-\tilde{y})\Big] \ .
\end{split}
\end{equation*}
This construction introduces no additional trainable parameters beyond those of $\breve{\mathbb{N}}^{(S_2)}_2(\tilde{x},\tilde{y})$; the constrained model is fully determined by the same parameter set through a fixed symmetrization operator. In other words, the neural parameters of $\breve{\mathbb{N}}^{(S_2)}_2(\tilde{x},\tilde{y})$ are equivalent to those of $\mathbb{N}^{(S_2)}_2(\tilde{x},\tilde{y})$.

The neural training procedure used to obtain an approximate \textit{normalized} Townes soliton employs loss contributions similar to those introduced in Section \ref{sec:fullpipe}, with the exception of the constraint that has already been ``hard'' pinned (always exactly satisfied). We aim to minimize the total loss function:
$$\mathcal{L}_{\mathbb{N}_2^{(S_2)},\mathbb{R}^{(S_2)},\mathbb{U}^{(S_2)}} = \sum^6_{j = 1} c^{(S_2)}_j \mathcal{L}^{(S_2)}_j \ , $$ 
in which the six contributions are:
\begin{itemize}
    \item The ``soft'' pinning conditions for the shoulders' values, which is the two-dimensional equivalence of Eq. \eqref{eq:Duffing_shoulder}:
    \begin{equation*}
    \begin{split}
        \mathcal{L}^{(S_2)}_1 &= \left[ \mathbb{N}_2^{(D)}(+0.5,0)- 1.0/3 \right]^2 \ , 
    \\
    \mathcal{L}^{(S_2)}_2 &= \left[ \mathbb{N}_2^{(D)}(-0.5,0)- 1.0/3 \right]^2 \ ,
    \\
    \mathcal{L}^{(S_2)}_3 &= \left[ \mathbb{N}_2^{(D)}(0,+0.5)- 1.0/3 \right]^2 \ ,
    \\
    \mathcal{L}^{(S_2)}_4 &= \left[ \mathbb{N}_2^{(D)}(0,-0.5)- 1.0/3 \right]^2 \ .
    \end{split}
    \end{equation*}
    \begin{widetext}
    \item The localization requirement, which defines the exterior region where the soliton fluctuation is small -- similar to \eqref{eq:Duffing_localization}:
    \begin{equation}
    \mathcal{L}^{(S_2)}_5 = \bar{E}\left\{ \left[ \text{ReLU}\left(|\mathbb{N}_2^{(D)}(\tilde{x}_j,\tilde{y}_j)|-1.0/3\right) \right]^2 \right\}_{\substack{
    |\tilde{x}_j| > 0.5 \ \lor \\
    |\tilde{y}_j| > 0.5
    }}\ ,
    \label{eq:loss_out_2D}
    \end{equation}
    where $\lor$ is the logical OR operation.
    \item The neural-learned SNL Schr\"odinger dynamics constraint, which is written in the proposed form Eq. \eqref{eq:SNL_Schr_1D_dyn_constraint}:
    \begin{equation}
        \mathcal{L}^{(S_2)}_6 = \bar{E}\left\{ \left[ \hat{X}^{(S_2)}_1(\tilde{x}_j,\tilde{y}_j)-\hat{X}^{(S_2)}_2(\tilde{x}_j,\tilde{y}_j) \right]^2 \right\}_{\forall j} \ ,
    \label{eq:Duffing_dyn_constraint}
    \end{equation}
    where the expressions $\hat{X}_1^{(S_2)}$ and $\hat{X}_2^{(S_2)}$ are:
    \begin{equation*}
        \begin{split}
            \hat{X}^{(S_2)}_1(\tilde{x}_j,\tilde{y}_j) = \mathbb{N}_1^{(S_1)}\left(\mathbb{U}^{(S_2)} \mathbb{N}_2^{(S_2)}(\tilde{x}_j,\tilde{y}_j) ,\frac{{\mathbb{U}^{(S_2)}}^2}{{\mathbb{R}^{(S_2)}}^2} \left\{ \left[\partial_{\tilde{x}}\mathbb{N}_2^{(S_2)}(\tilde{x}_j,\tilde{y}_j) \right]^2+\left[\partial_{\tilde{y}}\mathbb{N}_2^{(S_2)}(\tilde{x}_j,\tilde{y}_j) \right]^2\right\} \right) & \ ,
            \\
            \hat{X}^{(S_2)}_2(\tilde{x}_j,\tilde{y}_j) = \frac{\mathbb{U}^{(S_2)}}{{\mathbb{R}^{(S_2)}}^2} \left[ \partial_{\tilde{x}}^2 \mathbb{N}_2^{(S_2)}(\tilde{x}_j,\tilde{y}_j) + \partial_{\tilde{y}}^2 \mathbb{N}_2^{(S_2)}(\tilde{x}_j,\tilde{y}_j)\right]& \ ;
        \end{split}
    \end{equation*}
    these follows from the calculated expressions for the geometry-agnostic features in Eq. \eqref{eq:2D_feature}.
    \end{widetext}
\end{itemize} 
Note that we use the neural network model $\mathbb{N}_1^{(S_1)}$ obtained from the training for Eq. \eqref{eq:S1_train}. The weights are chosen to be $(c_1^{(S_2)},c_2^{(S_2)},c_3^{(S_2)},c_4^{(S_2)},c_5^{(S_2)},c_6^{(S_2)})=(1,1,1,1,1,25)$.

We compare the neural-found \textit{normalized} Townes solitary solution
$\mathbb{N}_2^{(D)}(\tilde{x},\tilde{y})$ with the high-precision numerical solution $\tilde{u}(\tilde{x},\tilde{y})$ --- obtained in Appedix \ref{app:2D_soliton_num} from knowing the exact governing PDE -- after training (see Fig. \ref{fig03}B1 and B2). The agreement between then are excellent, i.e. inside the training domain $(\tilde{x},\tilde{y}) \in [-\lambda,+\lambda] \times [-\lambda,+\lambda]$ where $\lambda = 5.0$, the relative root-mean-square ($L^2$-norm) error is low:
\begin{equation*}
    \frac{\bar{E}\left\{\left[ \mathbb{N}_2^{(D)}(\tilde{t}_j)     -  \tilde{u}(\tilde{t}_j) \right]^2\right\}_{\forall j}}{\bar{E} \left\{ \left[ \tilde{u}(\tilde{t}_j)\right]^2 \right\}_{\forall j}} \approx 0.02\% \ .
\end{equation*}
The neural-found rescaling factors are $\mathbb{R}^{(D)} \approx 2.68$ and $\mathbb{U}^{(D)} \approx 2.17$ also agrees with their numerical values $R^{(D)} \approx 2.61$ and $U^{(D)} \approx 2.21$. 

We have demonstrated that extreme scenarios can be identified in spaces of higher dimension than the one in which the dynamics were originally learned. We can also lift the learned dynamics to three-dimensional space and find that the neural-search still recovers the corresponding localized soliton profile with excellent agreement (see Fig. \ref{fig03}C) --- the relative root-mean-square ($L^2$-norm) is reasonably small $\approx 1.4\%$.

\subsection{Searching for a Liouville Instanton \label{sec:geoag}}

\begin{figure*}[!htbp]
\includegraphics[width=\textwidth]{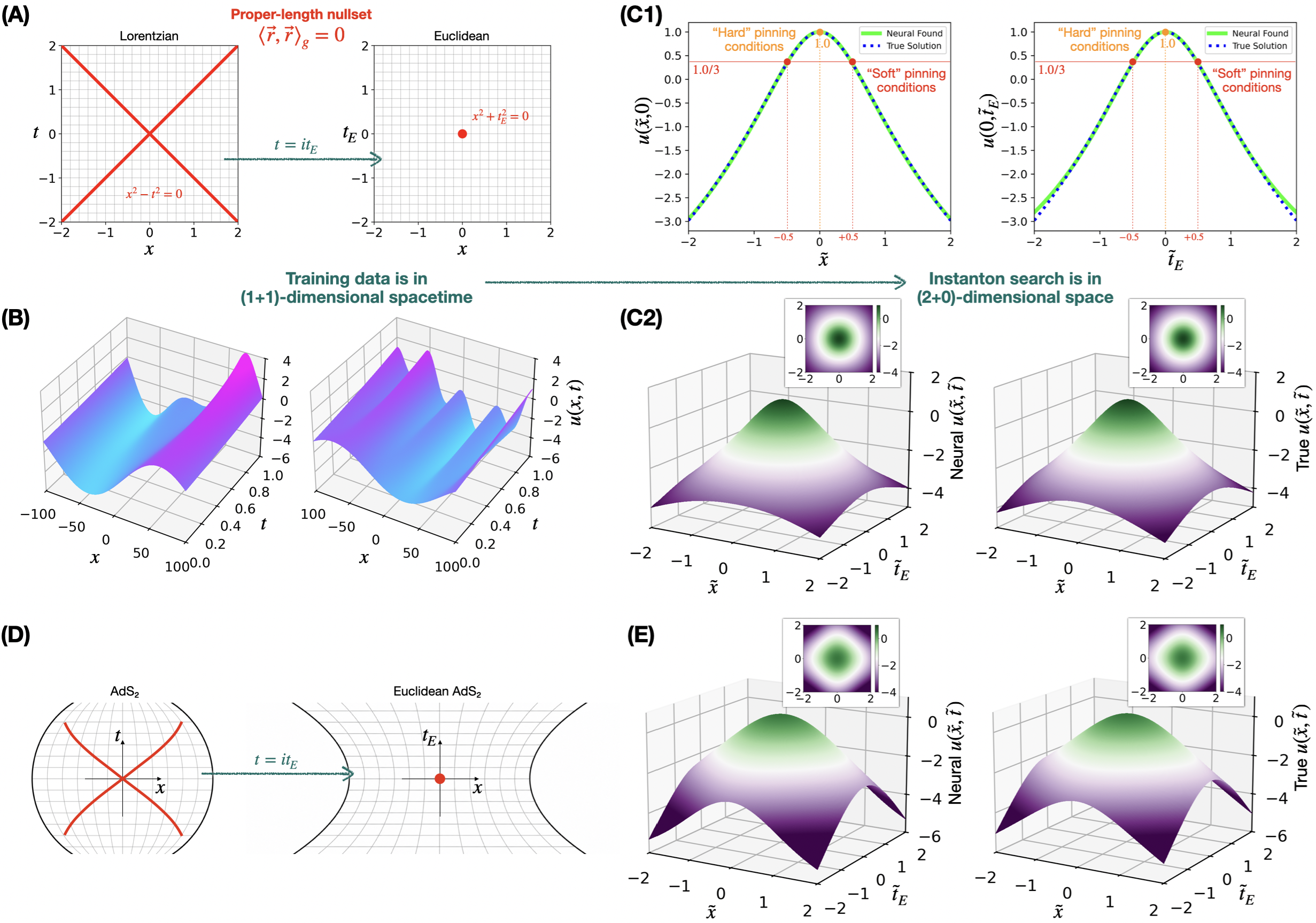}
\caption{\textbf{
Geometry-agnostic transfer from Lorentzian Liouville dynamics to a Euclidean
instanton search.}
The geometry-agnostic formation of the dynamics learned in real spacetime can be used,
after Wick rotation, to identify a localized instanton in imaginary time. \textbf{(A)} Lorentzian and Euclidean \textit{proper-length} nullsets.
\textbf{(B)} Representative field configurations used to train the dynamics-learning
neural network on a $(1+1)$-dimensional Lorentzian quasi-manifold.
\textbf{(C1)} Cross sections of the neural-found instanton along the $\tilde{x}$ and
$\tilde{t}_E$ axes agree with the analytical Liouville instanton profile.
\textbf{(C2)} Surface plots of the neural-found normalized instanton and the analytical
solution show good agreement over the Euclidean domain. \textbf{(D)} AdS$_2$ and Euclidean AdS$_2$ \textit{proper-length} nullsets. \textbf{(E)} Surface plots of the neural-found normalized instanton and the analytical
solution also show reasonable agreement over the Euclidean AdS$_2$ domain.}
\label{fig04}
\end{figure*}

For the final example, we consider a relativistic Liouville system in $(1+1)$-dimensional Lorentzian spacetime with coordinates $(x,t)$:
\begin{equation}
    \partial_t^2 u(x,t) - \partial_x^2 u(x,t)  = \exp\left[ u(x,t) \right] \ ,
    \label{eq:Liouville_quasiman}
\end{equation}
where $u(x,t)$ is a real scalar field. This PDE appears in many areas of theoretical
physics, including two-dimensional conformal field theory \cite{teschner2001liouville}, string theory \cite{polyakov1981quantum}, and
models of two-dimensional gravity \cite{distler1989conformal,david1988conformal}. In \underline{the first stage}, we learn the Liouville dynamics from data on a
Lorentzian quasi-manifold. In \underline{the second stage}, we Wick rotate it to a Euclidean
manifold and use the learned dynamics to search for an \textit{instanton}. It is important to note that Wick rotation \cite{wick1954properties} is one of the most powerful
analytic-continuation techniques in theoretical physics: by mapping the real-time $t$ to
an imaginary-time $t_E$, i.e. $t = it_E$ (see Fig. \ref{fig04}A for how such a change modifies the geometric structures), it turns certain real-time quantum
or field-theoretic processes into localized Euclidean configurations\footnote{It is important to note that, in general curved spacetime (especially in coordinate systems with mixed space-time terms such as Painlevé-Gullstrand coordinates \cite{painleve1921mecanique,gullstrand1922allgemeine}), a naive coordinate Wick rotation can lead to a complex metric; a more geometrically appropriate prescription is to Wick rotate the local orthonormal tetrad rather than the coordinate time itself \cite{samuel2016wick}.}. In this
formulation, tunneling events, saddle points, and other nonperturbative effects
can often be treated as mathematically well-posed solutions; their
collective contributions can dominate vacuum expectation values\footnote{This is realized explicitly in the Seiberg--Witten solution of
$\mathcal{N}=2$ supersymmetric Yang--Mills theory, where the nonperturbative
infrared dynamics can be determined exactly. In this setting, monopole
condensation produces confinement by the dual Meissner effect, showing that
Euclidean saddle sectors can determine the phase of a quantum field theory
rather than merely provide small perturbative corrections
\cite{seiberg1994electric,seiberg1994monopoles}.}  and other
physical observables \cite{coleman1979uses}. These solutions are
called instantons because they are localized in imaginary time, appearing as
coherent ``instant'' events rather than extended real-time processes.

In the flat Lorentzian $(1+1)$-dimensional spacetime:
\begin{equation}
    \begin{split}
    \langle \nabla_\Gamma u,\nabla_\Gamma u \rangle &= (\partial_x u)^2 - (\partial_t u)^2 \ ,
    \\
    \nabla_\Gamma^2 u &= \partial_x^2 u - \partial_t^2 u \ .
\end{split}
     \label{eq:1+1D_feature}.
\end{equation}
In the flat Euclidean $(2+0)$-dimensional space, we already have the calculation for the low-order geometry-agnostic features -- reported in Eq. \eqref{eq:2D_feature}:
\begin{equation}
    \begin{split}
    \langle \nabla_\Gamma u,\nabla_\Gamma u \rangle &= (\partial_x u)^2 + (\partial_{t_E} u)^2 \ ,
    \\
    \nabla_\Gamma^2 u &= \partial_x^2 u + \partial_{t_E}^2 u \ .
\end{split}
     \label{eq:2+0D_feature}.
\end{equation}

\ \ 

\subsubsection{The First Stage: \\ Dynamics-Learning from Data on a Quasi-Manifold} 

Following Appendix \ref{app:data_Liouville}, we generate a synthetic training dataset consisting of geometric-agnostic features as expressed in Eq. \eqref{eq:1+1D_feature}, i.e.
$$\{u_j,(\partial_x u_j)^2 - (\partial_t u_j)^2,\partial_x^2 u_j - \partial_t^2 u_j,...\}_{j=1}^{M_1^{(L_L)}}$$ from randomly-selected field configurations of
Eq. \eqref{eq:Liouville_quasiman} (see Fig. \ref{fig04}B). Since the underlying dynamics are not known \textit{a priori}, we assume a broad class of local dynamical laws by retaining all candidate terms up to a prescribed derivative order $\delta$, which we choose to be $\delta=2$. We then train the neural parameters of the
dynamics-learning neural network, denoted by $\mathbb{N}_1^{(L_L)}$, which
takes two-inputs
$$\Big(u_j,(\partial_x u_j)^2 - (\partial_t u_j)^2\Big)$$
and returns one-output
$$\mathbb{N}_1^{(L_L)}\Big(u_j,(\partial_x u_j)^2 - (\partial_t u_j)^2\Big) \ .$$ 
A visualization of this neural network can be found in Fig. \ref{figS01}A2. This training is done by minimizing
the squared-deviation loss function:
\begin{widetext}
    $$\mathcal{L}_{\mathbb{N}_1^{(L_L)}} = \bar{E}\left\{ \left[ \mathbb{N}_1^{(L_L)}\Big(u_j,(\partial_x u_j)^2 - (\partial_t u_j)^2\Big) - \left(\partial_x^2 u_j - \partial_t^2 u_j\right) \right]^2 \right\}_{\forall j} \ , $$
\end{widetext}
After training, $\mathbb{N}_1^{(D)}$ encodes the data-driven neural-learned dynamics of the Duffing system, i.e.
\begin{equation}
     \mathbb{N}_1^{(L_L)}\Big(u_j,(\partial_x u)^2 - (\partial_t u)^2\Big) \approx \left(\partial_x^2 u - \partial_t^2 u\right) \ .
    \label{eq:Duffing_learned}
\end{equation}

\subsubsection{The Second Stage: \\ Scenario-Searching on a Manifold} 

We use the previously trained network $\mathbb{N}^{(L_L)}_1$ to guide the
scenario-searching network $\mathbb{N}^{(L_E)}_2$. The superscripts $(L_L)$ and
$(L_E)$ indicate the two geometric settings: the dynamics-learning stage is
trained on data from a Lorentzian quasi-manifold, while the scenario-searching
stage is carried out on the corresponding (geometry-agnostic transferred) Euclidean manifold. A visualization of the neural network $\mathbb{N}^{(L_E)}_2$ can be found in Fig. \ref{figS01}B2. The search for a symmetric instanton is mostly identical to what has been done in Section \ref{sec:search_SNL_Schr} for a symmetric soliton; except that:
\begin{itemize}
    \item Unlike with the Duffing homoclinic pulse and the Townes solitons, here is no fixed peak value for the Liouville instanton, so we are free to choose the
    field value at its center. Without loss of generality, we impose
    $u(0,0)=1.0$ (but any other value is also fine). Then, the only trainable parameter, besides the
    neural parameters of $\mathbb{N}^{(L_E)}_2$, is $\mathbb{R}^{(L_E)}$, i.e. the
    common coordinate-rescaling factor for both $x$ and $t_E$. No field-rescaling means $\mathbb{U}^{(L_E)}=U^{(L_E)}=1$, and thus $\tilde{u}=u$.
    \item We do not require the field fluctuation around $0$ to remain small in the exterior region, but only
    require the exterior values to remain lower than the shoulder's values, i.e.
    $$u(\tilde{x},\tilde{t}_E)\Big|_{\substack{
    |\tilde{x}| > 0.5 \ \lor \\
    |\tilde{t}_E| > 0.5
    }} < u(0,0)/\eta \ \text{where} \ \eta=3 \ . $$
    Thus, instead of Eq. \eqref{eq:loss_out_2D}, we need no absolute-value operation:
    $$\mathcal{L}^{(L_E)}_5 = \bar{E}\left\{ \left[ \text{ReLU}\left(\mathbb{N}_2^{(L_E)}(\tilde{x}_j,\tilde{t}_{Ej})-1.0/3\right) \right]^2 \right\}_{\substack{
    |\tilde{x}_j| > 0.5 \ \lor \\
    |\tilde{y}_j| > 0.5
    }}\ , $$
    \item The weights $(c_1^{(L_E)},c_2^{(L_E)},c_3^{(L_E)},c_4^{(L_E)},c_5^{(L_E)},c_6^{(L_E)})$ are chosen to be $(25,25,25,25,1,1)$, so that the soft pinning conditions are enforced more strongly than the remaining loss terms.
\end{itemize} 

The Euclidean version of Eq. \eqref{eq:Liouville_quasiman} is given by:
\begin{equation}
    -\partial_{t_E}^2 u(x,t_E) - \partial_x^2 u(x,t_E)  =\exp\left[ u(x,t_E) \right] \ ,
    \label{eq:Euc_Lio}
\end{equation}
which admits an analytic symmetric instanton solution localized in
both space and imaginary-time (hence also called \textit{localized bump}):
\begin{equation}
    u(x,t) = \ln \left( \frac{8\gamma}{\left\{1 + \gamma\Big[(x-x_*)^2+(t-{t_E}_*)^2\Big]\right\}^{2}} \right) \ ,
    \label{eq:Lio_instanton}
\end{equation}
where $\gamma>0$ is an arbitrary positive constant, and
$(x_*,{t_E}_*)$ denotes the location at which the instanton emerges -- because of translational symmetry, this center can be anywhere.

After training $\mathbb{N}^{(L_E)}_2$ with locations randomly-selected as described in Appendix \ref{app:data_Liouville}, we compare the neural-found \textit{normalized} Liouville instanton
$\mathbb{N}_2^{(L_E)}(\tilde{x},\tilde{y})$ with the analytical solution $u(\tilde{x},\tilde{y})$ (see Fig. \ref{fig04}C1 and C2). The agreement between then are excellent, i.e. inside the whole training domain $(\tilde{x},\tilde{y}) \in [-\lambda,+\lambda] \times [-\lambda,+\lambda]$ where $\lambda = 2.0$, the relative root-mean-square ($L^2$-norm) error is reasonably low:
\begin{equation*}
    \frac{\bar{E}\left\{\left[ \mathbb{N}_2^{(L_E)}(\tilde{x}_j,\tilde{t}_{Ej})     -  \tilde{u}(\tilde{x}_j,\tilde{t}_{Ej}) \right]^2\right\}_{\forall j}}{\bar{E} \left\{ \left[ \tilde{u}(\tilde{x}_j,\tilde{t}_{Ej})\right]^2 \right\}_{\forall j}} \approx 0.9\% \ .
\end{equation*}
The neural-found rescaling factor $\mathbb{R}^{(L_E)} \approx 2.16$ is about the same with its true value $R^{(L_E)} \approx 2.16$. 

This final example demonstrates a successful knowledge transfer in our
geometry-agnostic framework that goes beyond what has been considered in previous
work \cite{phan2025towards}, while also addressing a task that arises naturally in many high-energy
physics systems. We can also apply the same procedure\footnote{Note that instead of a single trainable rescaling parameter for both $x$ and $t_E$, in general we need two of them. The training data set can also be enlarged to further improve the neural learned dynamics.} to more complex geometric structures, e.g. Euclideanized anti-de Sitter (AdS$_2$) space\footnote{The proper line-element in AdS$_2$ spacetime is given by: $$ds^2=a^2(t)dx^2-dt^2 \ , $$
where the scale-factor is $a(t)=\cos(t)$; this spacetime has a constant negative Ricci scalar curvature $2\ddot{a}(t)/a(t)=-2$. After Wick rotation, we arrive at:
$$ds^2=\cosh^2(t_E) dx^2 + dt_E^2 \ . $$
This is Euclidean AdS$_2$ space, which is also equivalent to an hyperbolic plane. In there, the first few low-order geometry-agnostic features can be calculated:
\begin{equation*}
\begin{split}
    \langle \nabla_\Gamma u,\nabla_\Gamma u \rangle = \frac{(\partial_x u)^2}{\cosh^2(t_E)} + (\partial_{t_E} u)^2 & \ ,
    \\
    \nabla_\Gamma^2 u = \frac{\partial_x^2 u}{\cosh^2(t_E)} + \tanh(t_E) \partial_{t_E} u+ \partial_{t_E}^2 u & \ .
\end{split}
\end{equation*}} \cite{henneaux1985asymptotically,strominger1999ads} (see Fig. \ref{fig04}D), and find that the neural-search still approximate the corresponding instanton (known analytically if including the scalar curvature term to the dynamics) with reasonable precision (see Fig. \ref{fig04}E) --- the relative root-mean-square ($L^2$-norm) error is not large $\approx 11\%$.

\section{Discussion \& Conclusion}

We have proposed a two-stage, geometry-agnostic, data-driven framework for
identifying plausible extreme scenarios. A dynamics-learning neural network
first infers the governing constraint from observational data. A
scenario-searching neural network then uses this learned constraint as a
physics-informed residual to identify coherent structures that are unlikely to
appear in the original data, may arise in a new physical setting, or may even
belong to an analytically continued setting that is not directly physical but is
theoretically important. In all considered examples, the agreement is very strong, supporting the
feasibility of our framework.

Two related search problems should be distinguished. At fixed governing equation parameters, one can seek initial conditions or trajectories in phase space that produce an extreme excursion. Alternatively, one can search parameter space for values at which a target structure becomes possible, such as a homoclinic orbit in a dissipative system. Our examples address structure discovery at fixed governing equation parameters; the optimized rescaling factors determine the candidate profile’s physical scales within those fixed dynamics. Extending the framework to parameter discovery would require a dynamics-learning model conditioned on the physical parameters, supported by data across the relevant parameter range, and a joint search over those parameters and the candidate solution.

Parameter space searches also connect naturally to tipping-point analysis. Near certain bifurcations, changes in stability can increase susceptibility to abrupt, fluctuation-driven transitions. Fabiani et al. \cite{fabiani2024task} has demonstrated how learned surrogate models can locate tipping points and quantify rare-event escape times near them. Their work suggests a complementary extension of our framework: combining coherent-structure searches with bifurcation and stability analysis to identify parameter regimes in which extreme transitions become accessible or more likely.

Several limitations remain. The present examples use synthetic data without
artificial noise, while realistic applications will involve noisy measurements,
sparse sampling, hidden variables, boundary effects, and model mismatch. Future
work should therefore combine the present framework with uncertainty
quantification, active data acquisition, symmetry-aware architectures, and
stability analysis of the discovered solutions. A natural next step is to apply
the framework to \textit{experimentally accessible} active-matter systems, such as
chemotactic bacteria, where both the bacterial density and the self-generated
chemical field can be measured dynamically \cite{phan2024direct}. In such a
setting, one could train the first stage network to learn an effective
chemotactic field equation from image data of wave experiments in one-dimensional channel \cite{phan2024direct}, and then use the second stage search
to ask whether the learned dynamics support localized coherent structures in two-dimensional space, such
as a self-sustained bacterial-density bump maintained by the chemical field
generated by the bacteria themselves, in either static or dynamical enviroment settings \cite{phan2020bacterial,phan2024social}. As a concrete invitation, we note that Eq. \eqref{eq:Euc_Lio} also arises in a simple limiting case of an \textit{E. coli} chemotaxis model \cite{brenner1998physical}:
\begin{equation*}
\begin{split}
    \partial_t \rho(\vec{r},t) & \approx \vec{\nabla} \left[ \rho(\vec{r},t) \vec{\nabla}c(\vec{r},t) \right] \ ,
    \\
    0 &\approx \vec{\nabla}^2 c(\vec{r},t) + \rho(\vec{r},t) \ ,
\end{split}
\end{equation*}
where $\rho(\vec{r},t)$ is the bacteria density and $c(\vec{r},t)$ is the concentration of the chemical secreted by the bacteria themselves. From the first equation, in the quasi-stationary limit, i.e. $\partial_t \rho \approx 0$, the term $\rho \vec{\nabla}c$ is approximately divergence-free, thus leading to the
Boltzmann-type balance $\rho \propto \exp(c)$; substituting this relation into the second equation then yields:
\begin{equation}
    -\vec{\nabla}^2 c(\vec{r}) \propto \exp\left[c(\vec{r})\right] \ ,
\end{equation}
which is identical to \eqref{eq:Euc_Lio} (up to relabeling and field-rescaling). The Liouville instanton corresponds to a stationary bacterial aggregation, a phenomenon that has already been observed experimentally \cite{budrene1991complex}.

There are immediate directions for extending this work, on the \textit{theoretical side}. One natural
next step is to move beyond scalar fields. Fermionic fields provide a
particularly interesting target because spinor dynamics depend directly on the
spin connection, and torsion can couple to fermionic degrees of freedom in a
physically meaningful way \cite{lawson2016spin,hehl1976general,shapiro2002physical}. In such systems, the geometric data are not merely
background ingredients for scalar Laplacians; they enter the Dirac operator
itself. A geometry-agnostic learning framework for fermionic fields would
therefore need to learn from spinor-valued features, covariant derivatives, and
connection-dependent invariants. This would provide a direct test of whether
the present approach can handle settings in which torsion is not a small
geometric complication, but a central part of the field dynamics, and what extreme scenarios can emerge from those contributions of non-Riemannian geometric structures.

Another extension is to go beyond geometry-agnostic learning for individual fields and consider collections of fields tied together by internal symmetries. Supersymmetric field theories provide a natural setting for this direction, because bosonic and fermionic fields are organized into the same supermultiplet \cite{wess2020supersymmetry,weinberg2005quantum,freedman2012supergravity}. There, the dynamical constraints should not be learned separately for each component field: doing so may break the supersymmetry relations that make the theory consistent. Instead, the framework should learn coupled constraints across the entire multiplet, using geometric features for each field while also respecting the symmetry transformations that relate them. Moreover, supersymmetric theories often provide exact or semi-exact access to
nonperturbative phenomena, including instanton and monopole solutions \cite{seiberg1994electric,seiberg1994monopoles}, thus
providing another playground where extreme scenarios may appear more
ubiquitously and can be checked against analytically know-results.

\ \ 

\textbf{AI Transparency Statement:} OpenAI GPT-6 models accessed through Codex assisted with initial drafting, language polishing, literature search, and coding for figure generation and manuscript preparation. Parts of the illustrations (cartoons) in Fig. \ref{fig01} and Fig. \ref{figS01} are AI generated. The ideas, claims, and framing are the authors'. The authors assume responsibility for all content.

\bibliography{main}
\bibliographystyle{apsrev4-2}

\appendix

\section{Training Data and Locations \label{app:data}}

\subsection{Duffing ODE \label{app:data_Duffing}}

Training data were generated from $\ddot u=u-u^3$ (Duffing) and
$\ddot u=-u$ (simple harmonic oscillator, SHO). For each system,
$10^3$ trajectories were integrated over $t\in[0,10]$ using classical
fourth-order Runge--Kutta with $\Delta t=10^{-3}$. Initial positions
and velocities were sampled independently and uniformly from $[-2,2]$
(NumPy seed $37$). Centered finite differences of the simulated positions
provided $\dot u$ and $\ddot u$ at the $9999$ interior time points of each
trajectory. Sampling $0.2\%$ of the pooled points without replacement
produced $\sim 2 \times 10^4$ input--target pairs $(u,\dot u)\rightarrow \ddot u$
per system. No measurement noise was added. Scenario-searches used $10^3$ uniformly sampled collocation points on
$\tilde t\in[-5,5]$. The central region was $|\tilde t|<0.5$, and
shoulder constraints were imposed at $\tilde t=\pm0.5$. Collocation
points were sampled once and held fixed during training.

\subsection{SNL Schr\"odinger PDE \label{app:data_Schr}}

A separate dynamics network was trained on the same sampled Duffing
trajectories using the squared-derivative input,
$\big(u,{u'}^2 \big)\rightarrow u''$. This representation was transferred to the
SNL Schr\"odinger equation by replacing ${u'}^2$
with $\big( \vec{\nabla} u \big)^2$ and $u''$ with $\vec{\nabla}^2 u$. No additional
higher-dimensional simulation data were used to train this network. The one-dimensional search used the collocation scheme described in
Appendix \ref{app:data_Duffing}. 

For dimensions $d=2,3$, the search
domain was $[-5,5]^d$. Each search used $10^4$ uniform collocation
points inside $[-0.5,0.5]^d$ and $10^4$ outside this inner box but within
the search domain, obtained by rejection sampling. These points were
sampled once, and the shoulder constraints were imposed at the $2d$ points located distance $0.5$ away from the origin, along each positive and negative coordinate axis.

\subsection{Liouville PDE \label{app:data_Liouville}}

The dynamics was learned in flat Lorentzian spacetime. We simulated Eq. \eqref{eq:Liouville_quasiman} on the periodic domain
$x\in[-100,100)$ with $\Delta x=1$, over $t\in[0,1]$ with
$\Delta t=10^{-5}$. Spatial derivatives used centered differences,
and time integration used classical fourth-order Runge--Kutta.
The initial conditions were
\begin{align}
 u(x,0)&=-2+A\cos(n\pi x/100+\phi),\\
 \partial_t u(x,0)&=B\sin(m\pi x/100+\psi),
\end{align}
where the coefficients $A,B$ were uniform on $[1,3]$, the integers $n,m$ were randomly selected from
$\{1,2,3,4\}$, and the phases $\phi,\psi$ were uniform on $[0,2\pi]$. Runs with field-value exceeding $u=10$ were rejected. The recorded
dataset contains $19$ accepted runs out of $20$ attempts, with $10^3$
spacetime points sampled without replacement from each accepted run.
Inputs were 
$$\left(u,(\partial_t u)^2 - (\partial_x u)^2 \right) \ , $$
with $\partial_t u$ taken from the evolved velocity
field. Target constraints Eq. \eqref{eq:Liouville_quasiman} were evaluated using the simulator
right-hand side. Further reducing sample size with field-value $u<2$ retains $\sim 1.8 \times 10^4$ pairs.

For both the Euclidean flat and AdS$_2$ searches, the dimensionless domain was
$[-2,2]^2$. We used $10^4$ collocation points inside $[-0.5,0.5]^2$
and $10^4$ outside it, sampled and held fixed as in the SNL searches.
The $4$ shoulder locations were at distance $0.5$ away from the origin, along each positive and negative coordinate axis.

\section{Neural Architecture \label{app:NN}}

\begin{figure*}[!htbp]
\includegraphics[width=0.8\textwidth]{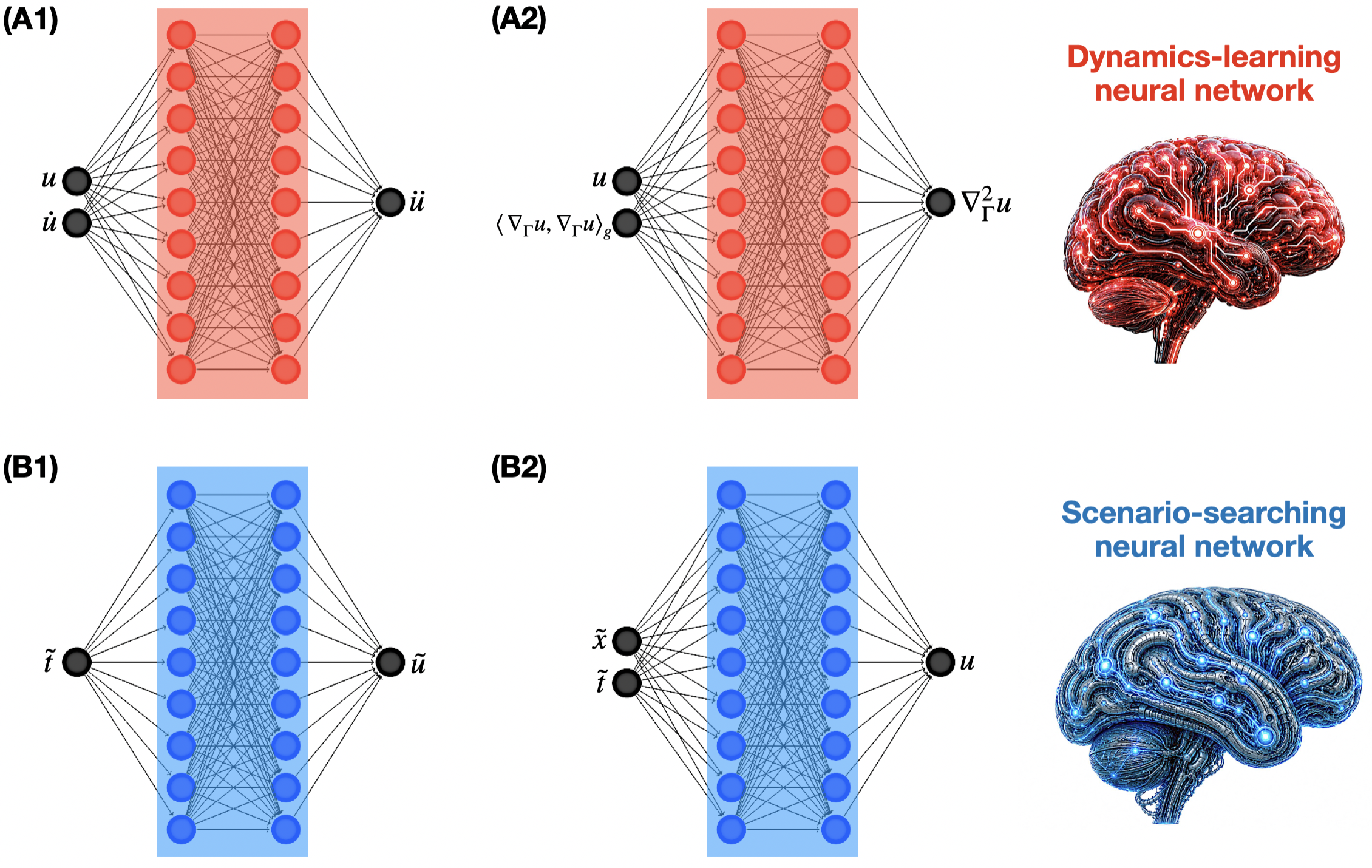}
\caption{\textbf{Neural network architectures used in this work.} All architectures are standard feedforward MLPs.}
\label{figS01}
\end{figure*}

Two networks serve distinct roles. The dynamics-learning network maps
local field features to the differential quantity learned from the data (see Fig. \ref{figS01}A).
The scenario-search network maps dimensionless coordinates to
a candidate profile, whose derivatives are evaluated by automatic
differentiation (see Fig. \ref{figS01}B). Its parameters are optimized together with physical
rescaling parameters, using the learned dynamics and the desired
scenario constraints. All networks are fully connected MLPs with $\tanh$ hidden activations
and a linear raw output. Each dynamics network has two inputs, one
hidden layer of width $64$, and one output. Search networks have $d$
coordinate inputs, one output, and the hidden widths listed in
Table \ref{tab:implementation}.

\begin{table*}[t]
\centering
\small
\setlength{\tabcolsep}{10pt}
\begin{tabular}{lccccc}
\hline
Search & Hidden widths & Collocation points & Epochs & $\eta_\theta$ & $\eta_s$ \\
\hline
Duffing / SHO & $32$ & $10^3$ & $5 \times 10^4$ & $10^{-4}$ & $10^{-4}$ \\
SNL, 1D & $32$ & $10^3$ & $5 \times 10^4$ & $10^{-4}$ & $10^{-4}$ \\
SNL, 2D & $32,32$ & $2\times10^4$ & $5 \times 10^3$ & $10^{-3}$ & $10^{-2}$ \\
SNL, 3D & $32,32,32$ & $2\times10^4$ & $5 \times 10^3$ & $10^{-2}$ & $10^{-1}$ \\
Liouville, Euclidean flat & $32,32$ & $2\times10^4$ & $1.5 \times 10^4$ & $10^{-3}$ & $10^{-2}$ \\
Liouville, EAdS$_2$ & $32,32,32$ & $2\times10^4$ & $3 \times 10^3$ & $10^{-3}$ & $10^{-3}$ \\
\hline
\end{tabular}
\caption{Scenario-search settings. Hidden widths exclude input and output
layers. $\eta_\theta$ and $\eta_s$ denote learning rates for network and
rescaling parameters, respectively. }
\label{tab:implementation}
\end{table*}

The one-dimensional searches used the raw network output. For the SNL Schr\"{o}dinger and Liouville systems searches, coordinate-reflection symmetry was enforced by averaging the network output over all coordinate reflections and normalizing it by its value at the origin. The result was then scaled to set the center value to $1$ for SNL Schr\"{o}dinger systems and $1/2$ for Liouville systems. This construction enforces reflection symmetry along each coordinate, but does not impose full rotational symmetry.

\section{Neural Training Procedures \label{app:training}}

For dynamics learning, each input component and the target were divided
by their respective maximum absolute values over the retained dataset,
before a random $90:10$ pointwise training--validation split.
Training minimized mean squared error using Adam and shuffled batches
of $1024$ samples. The Duffing, SHO, and squared-derivative Duffing
models used $2\times 10^3$ epochs at learning rate $5\times10^{-4}$;
the Liouville model used $5 \times 10^4$ epochs at $10^{-4}$.
The final model was retained rather than selected by early stopping.

During scenario searches, the dynamics-network weights were excluded
from the optimizer. Full-batch Adam updated the search-network weights
and rescaling parameters using the settings in
Table \ref{tab:implementation}. Positive scales were parametrized as
$\operatorname{softplus}(a)+10^{-6}$. In one dimension, the raw scale
parameters started at $a=1$, corresponding to scales approximately
$1.31$; in the higher-dimensional searches, scales started at
approximately $1$. No reference-profile samples entered the training loss.

It is important to note that the two Liouville searches (Fig. \ref{fig04}C2 and Fig. \ref{fig04}E) required additional
optimization to improve convergence. For these cases, the Adam learning
rates were reduced to $10^{-6}$ by cosine annealing, followed by
300 iterations of full-batch L-BFGS (with learning rate $1$, history
size $50$, and a strong-Wolfe line search). This refinement optimized
the network weights and rescaling parameters (i.e. $L$ for the flat search and
$R,T$ for EAdS$_2$), using the same loss and collocation points.

\section{Generate Numerical Profile \\ of Townes Soliton \label{app:2D_soliton_num}}

The radially symmetric 2D reference profile was computed from
\begin{equation}
 \begin{gathered}
 Q''(r)+\frac{1}{r}Q'(r)=Q(r)-Q(r)^3,\\
 Q'(0)=0,\qquad Q(r)\to0\quad(r\to\infty).
 \end{gathered}
\end{equation}
We solved the truncated boundary-value problem on $[10^{-6},30]$
with $Q'(10^{-6})=0$ and $Q(30)=0$, using SciPy's
\texttt{solve\_bvp}. The initial mesh contained $5 \times 10^3$ uniformly spaced
points, with initial guesses $Q=2.2e^{-r}$ and $Q'=-2.2e^{-r}$.
The solver tolerance was $10^{-6}$, with at most $5 \times 10^4$ mesh nodes. The converged profile was tabulated at $5000$ radii on $[0,30]$ and
linearly interpolated, taking $Q(0)\rightarrow Q(10^{-6})$. To match the
scenario constraints, we set
\begin{equation}
 U_*=Q(0),\qquad R_*=2r_{1/3},\qquad
 v_*(\rho)=\frac{Q(R_*\rho)}{U_*},
\end{equation}
where $Q(r_{1/3})=Q(0)/3$. This gives $v_*(0)=1$,
$v_*(0.5)=1/3$, $U_*\simeq2.21$, and $R_*\simeq2.61$. 

For the 3D reference, the radial term becomes $2Q'/r$. Shooting on
$[10^{-6},25]$ used the regular-origin expansion
$Q=A+(A-A^3)r^2/6$, $Q'=(A-A^3)r/3$, and the outer condition
$Q'(25)+(1+1/25)Q(25)=0$. SciPy's \texttt{solve\_ivp} used
relative/absolute tolerances $10^{-8}/10^{-10}$ and maximum step
$0.05$; \texttt{brentq} searched $A\in[4,4.5]$ with absolute tolerance
$10^{-10}$. The same tabulation and normalization procedure yielded
$U_*\simeq4.34$ and $R_*\simeq1.55$.

\end{document}